\documentclass{article}

\usepackage{microtype}
\usepackage{graphicx}
\usepackage{subfigure}
\usepackage{booktabs} 
\usepackage{graphicx}
\usepackage{times}
\usepackage{soul}
\usepackage{url}
\usepackage[hidelinks]{hyperref}
\usepackage[utf8]{inputenc}
\usepackage{graphicx}
\usepackage{amsmath}
\usepackage{booktabs}
\usepackage{subfigure,tabularx,algorithmic,units,hyperref}
\usepackage{amsfonts}
\usepackage{bbm}
\newtheorem{definition}{Definition}

\usepackage{hyperref}
\usepackage{amsmath}

\usepackage[accepted]{hill2019}

\icmltitlerunning{Issues with post-hoc counterfactual explanations: a discussion}

\begin{document}

\twocolumn[
\icmltitle{Issues with post-hoc counterfactual explanations: a discussion}



\icmlsetsymbol{equal}{*}

\begin{icmlauthorlist}
\icmlauthor{Thibault Laugel}{lip6}
\icmlauthor{Marie-Jeanne Lesot}{lip6}
\icmlauthor{Christophe Marsala}{lip6}
\icmlauthor{Marcin Detyniecki}{lip6,axa,pol}
\end{icmlauthorlist}

\icmlaffiliation{lip6}{Sorbonne Universit\'e,
CNRS, Laboratoire d’Informatique de Paris 6, LIP6, F-75005}
\icmlaffiliation{pol}{Polish Academy of Science, IBS PAN, Warsaw, Poland}
\icmlaffiliation{axa}{AXA, Paris, France}

\icmlcorrespondingauthor{Thibault Laugel}{thibault.laugel@lip6.fr}

\icmlkeywords{Machine Learning interpretability, ICML}

\vskip 0.3in
]



\printAffiliationsAndNotice{}

\begin{abstract}

Counterfactual post-hoc interpretability approaches have been proven to be useful tools to generate explanations for the predictions of a trained blackbox classifier. However, the assumptions they make about the data and the classifier make them unreliable in many contexts. In this paper, we discuss three desirable properties and approaches to quantify them: proximity, connectedness and stability. In addition, we illustrate that there is a risk for post-hoc counterfactual approaches to not satisfy these properties.

\end{abstract}

\section{Introduction}
\label{sec:introduction}

Among the soaring number of methods proposed to generate explanations for classifiers, post-hoc interpretability aproaches~\cite{Guidotti2018survey} have been the subject of debates recently in the community~\cite{Rudin2018}. By generating explanations for the predictions of a trained predictive model without using any knowledge about it whatsoever (i.e. treating it as a \textit{blackbox}), these systems are inherently flexible enough to be used in any situation~(model, task...) by any user, which makes them popular today in various industries. However, their main downside is that, under these assumptions, there is no guarantee that the built explanations  are faithful to the original data that were used to train the model. 


This question especially applies to counterfactual example approaches (e.g.~\citet{Martens2014,Wachter2018,Guidotti2018lore,Russell2019} that, based on counterfactual reasoning (see e.g.~\citet{Bottou2013}), aim at answering the question: \textit{given a trained classifier and an observation, how is its prediction altered when the observation changes?}
In the context of classification, they identify the minimal perturbation required to change the predicted class of a given observation: a user is thus able to understand what features locally impact the prediction and therefore how it can be changed.
Among interpretability methods, counterfactual examples have been shown to be useful solutions~\cite{Wachter2018} that can be easily understood and thus directly facilitate a user's decisions.

However, without any knowledge on ground-truth data nor on the classifier, counterfactual examples in the post-hoc paradigm are vulnerable to issues raised by the robustness and complexity of the classifier (such as overfitting or excessive generalization), leading to explanations that are not satisfying in the context of interpretability.

This work proposes a discussion over three properties that we argue a counterfactual example should satisfy to design useful explanations: proximity, connectedness and stability. This paper aims to motivate these properties and discuss approaches to quantify them. In addition, we illustrate that, in the post-hoc context, there is a risk for counterfactual example approaches of generating explanations that do not respect these criteria, leading to misleading or useless explanations.

In Section~\ref{sec:background} of this paper, a brief overview of counterfactual approaches is presented with a focus on the post-hoc context. Sections~\ref{sec:proximity}, \ref{sec:connectedness} and~\ref{sec:stability} are devoted to presenting and motivating these three properties. Finally, a discussion is proposed in Section~\ref{sec:discussion}.

\section{Background}
\label{sec:background}

\subsection{Counterfactual Examples}

Instead of simply identifying important features (for the model) like most interpretability approaches~(e.g. SHAP~\cite{Ljundberg2017}), counterfactual example approaches aim at finding the minimal perturbation required to alter a given prediction. A counterfactual explanation is thus a specific data instance, close to the observation whose prediction is being explained, but predicted to belong to a different class.
This form of explanation provides a user with tangible explanations that are directly understandable and actionable as it answers a natural question raised by explanations: \textit{"Would changing a certain factor have changed the decision?"}~\cite{Doshi-Velez2018}. This can be opposed to feature importance vectors, which are arguably harder to use and to understand for a non-expert user~\cite{Wachter2018}.
Several formalizations of the counterfactual problem can be found in the literature, depending on the formulation of the minimization problem and on the used distance metric.
For instance, GS~\cite{Laugel2017inverse} (resp.~\citet{Guidotti2018lore}) look for the $L_2$ (resp.~$L_0$)-closest instance of an other class, while HCLS from~\citet{Lash2017} aims to find the instance with the highest probability of belonging to another class within a certain maximum distance. Another example is the problem in~\citet{Wachter2018} (and its adaptation to the post-hoc context by~\citet{McGrath2018}), formulated as a trade-off between the $L_1$ closest instance and a specific classification score target.

\subsection{Studies of Post-hoc Interpretability Approaches}

However, the post-hoc paradigm, and in particular the need for post-hoc approaches to use instances that were not used to train the model to build their explanations, raises questions about their relevance and usefulness. Troublesome issues have been in the context of non-conterfactual approaches: for instance, it has been noticed~\cite{Baehrens2010} that modeling the decision function of a black-box classifier with a surrogate model trained on generated instances can result in explanation vectors that point in the wrong directions in some areas of the feature space in trivial problems. The stability of post-hoc explainer systems has been criticized as well, showing that some of these approaches are locally not stable enough~\cite{Alvarez2018} or on the contrary too stable and thus not locally accurate enough~\cite{Laugel2018}. 

In line with some of these previous works, we discuss desiderata we believe a counterfactual explanation should satisfy. These three properties are defined in the three following sections.

\textbf{Notations}
In the rest of the paper, we note $f:\mathcal{X}\rightarrow \mathcal{Y}$ a classifier trained on the dataset $X$. Let $x$ be an instance of $\mathcal{X}$ whose prediction $f(x)$ we want to interpret with the counterfactual explainer $E$. $E(x)$ denotes a counterfactual example and as such belongs to $\mathcal{X}$ and satisfies, by construction, $f(x) \neq f(E(x))$. Let $d$ denote a distance function considered by the approach $E$ (e.g. $L_0$ for~\citet{Guidotti2018lore}). Additionally, we denote $X^l$ the set of instances of~$X$ correctly predicted to belong to class $l\in \mathcal{Y}$.

\section{Proximity}
\label{sec:proximity}

\subsection{Presentation }

\textbf{Notion }
The most intuitive desiderata for a counterfactual explanation is that it should provide a user with plausible means of actions to alter a prediction: for instance, telling a customer asking for a credit that his/her credit is rejected because he/she needs to earn a negative amount of money does not make any sense. We define this notion of plausibility using a distance and argue that a counterfactual should be close to an instance from ground-truth data from the same class in order to be useful to a user: the explanation~$E(x)$ is plausible because it looks like existing ground-truth knowledge.

\textbf{Proposition of Criterion } We thus propose to evaluate the distance between $E(x)$ and its closest neighbor~$a_0$ from~$X$ correctly predicted to belong to the same class: $a_0 \in X^{f(E(x))}$: having at least one instance should be enough to guarantee plausibility.
In order to have a relative metric, we propose to compare this distance to the distance between between $a_0$ and its closest neighbor from~$X^{f(E(x))}$: in order to be plausible, $E(x)$ should be approximately at the same distance from its closest neighbor than the latter is from the rest of the data.

Formally, $a_0=\underset{x_i \in X^{f(E(x))}}{\text{argmin}} d(E(x), x_i)$ and we propose to consider:
$$
P(E(x)) = \frac{d(E(x), a_0)}{\underset{x_i \in X^{f(E(x))}}{\text{min}} d(a_0, x_i)} 
$$
Note that this corresponds to the Local Outlier Factor score~\cite{Breunig2000}, used for outlier detection, with $k=1$. Indeed, the goal is to identify if a generated counterfactual example $E(x)$ is outlying with regard to ground-truth instances of the same class to ensure it is not an exception and can therefore be understood by the user.

\subsection{Illustrative Examples}
\begin{figure}[t]
\centering     
\includegraphics[width=0.75\columnwidth]
{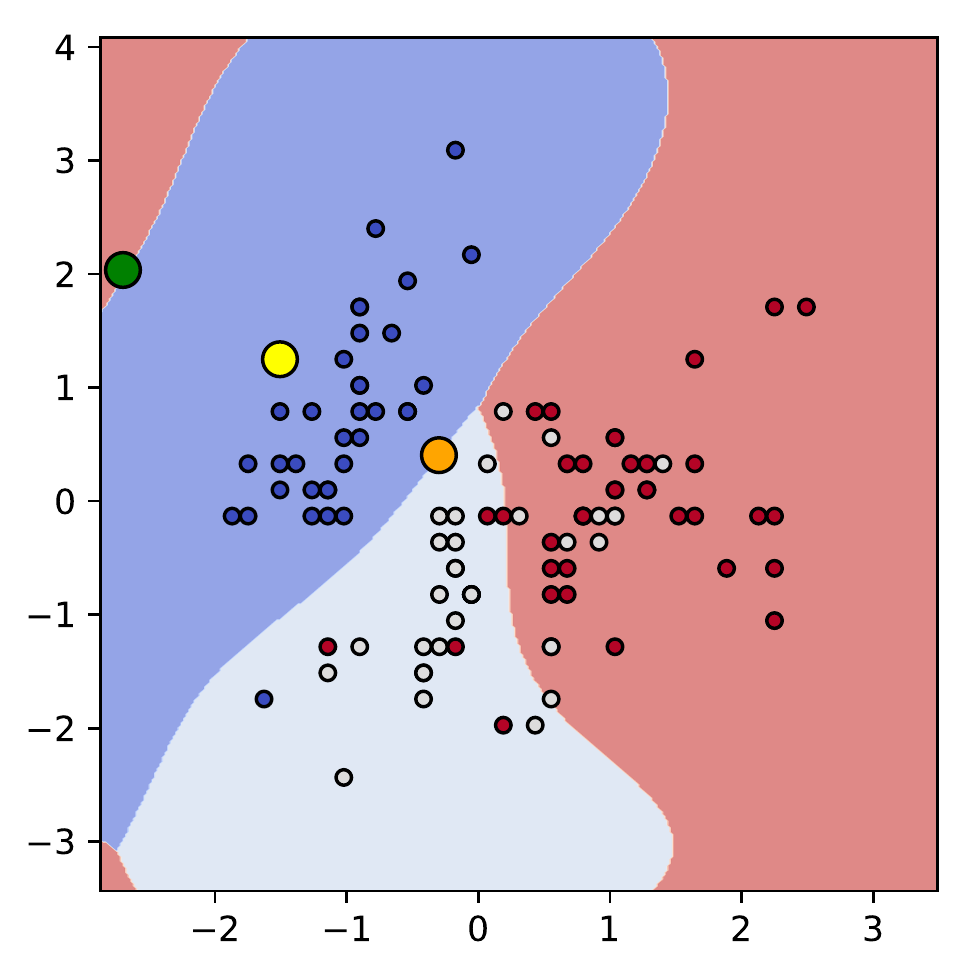}
\caption{Illustration of the intuition behind the Proximity metric for two counterfactual explanations generated using HCLS (green instance) and GS (orange instance), for a random instance (yellow) of the iris dataset.}
\label{fig:illustration-proximity}
\end{figure}

For the purpose of giving insights about the criterion $P$, a 2D version of the iris dataset is considered. A classifier (SVM classifier with RBF kernel and default scikit-learn parameters), is trained on $70\%$ of the data ($80\%$ accuracy on the rest of the dataset). Figure~\ref{fig:illustration-proximity} shows examples of what the intuition behind the proximity criterion as well as issues that can arise in the post-hoc context for a specific instance $x$ (yellow). The training instances and the learned decision boundaries of $f$ are represented by the colored (light blue, blue and red) instances and areas.
Two post-hoc counterfactual approaches, HCLS~\cite{Lash2017} and GS~\cite{Laugel2017inverse}, are used to generate counterfactual explanations to instances from the test dataset. In this case, the counterfactual generated by GS (orange instance) is located at a reasonable distance from training instances classified similarly: $P(E(x))=0.82$. However, the explanation generated using HCLS (green instance) is located far away from the training instances of its class: in this case, $P(E(x))=12.98$. Calculating this value for HCLS for all instances of the test set shows that while most instances ($86.7\%$) have a proximity score between $0$ and $3.0$, other have extremely high proximity scores (approx.~$14$). This confirms that in some cases, post-hoc counterfactual approach do indeed generate instances that cannot be associated to ground-truth data through distance.

\section{Connectedness}
\label{sec:connectedness}

\subsection{Presentation}
\textbf{Notion }
Another notion of ground-truth justification aims at favoring explanations that result of previous knowledge and ones that would not be a consequence of an artifact of the classifier. Artifacts can be created in particular because of a lack of robustness of the model $f$, leading to questionable "improvisations" in regions it has no information about (no training data). Although not harmful in the context of prediction (a desirable property of a classifier remains its ability to generalize to new observations), having an explanation caused by an artifact that cannot be associated to any existing knowledge by a human user may be undesirable in the context of interpretability.

We propose to define this relation between an explanation and some existing knowledge (ground-truth data used to train the blackbox model) using the topological notion of path. In order to be more easily understood and employed by a user, we argue that the counterfactual instance should be \textit{continuously} connected to an observation from the same class. This property is thus complementary to the Proximity one, as two instances can be close but not linked by a continuous path.
To adapt this notion to a blackbox classifier, we approximate this continuous notion with $\epsilon$-\textit{chainability} (with $\epsilon>0$) between two instances $e$ and $a$, meaning a finite sequence $e_0$, $e_1$, ... $e_N$ $\in \mathcal{X}$ exists such that $e_0=e$, $e_N=a$ and $\forall{i}< N,\, d(e_i, e_{i+1})<\epsilon$. This leads to the following definition:

\begin{definition}[$\epsilon$-connectedness] 
An instance $e \in \mathcal{X}$ is \emph{$\epsilon$-connected} to an instance $a \in \mathcal{X}$ if $f(e) = f(a)$ and if there exists an $\epsilon$-chain $(e_i)_{i<N} \in \mathcal{X}^N$ between $e$ and $a$ such that $\forall n<N, f(e_i)=f(e)$.
\label{def:epsilon-justification}
\end{definition}

\begin{figure}[t]
\centering     
\includegraphics[width=0.95\columnwidth]
{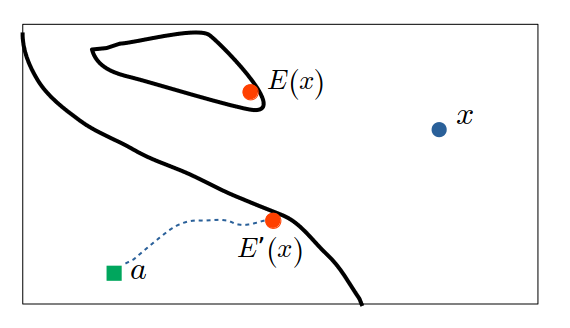}
\caption{Illustration of the intuition behind the notion of Connectedness.}
\label{fig:dessin-connectedness}
\end{figure}

Following this definition, an explanation $E(x)$ should be $\epsilon$-connected to some ground-truth instance. The idea behind this connectedness notion is to identify the instances from the training data that are being predicted to belong to the same class for similar reasons. An illustration of the notion of connectedness is shown in Figure~\ref{fig:dessin-connectedness} in two dimensions for a binary classifier (black decision border): two counterfactual explanations are generated (orange instances) for an instance $x$ (blue). One of them, $E(x)$ lies in a classification region that does not contain any training instance, while the other, $E'(x)$, can be connected with the region to $a\in X$.

This notion is further studied in~\cite{Laugel2019}

The connectedness of classification regions has been studied in the contest of deep neural networks for image classification (e.g.~\citet{Fawzi2018}), but not in the context of interpretability. 

\textbf{Proposition of Criterion }
We thus propose to assess the connectedness of an explanation $E(x)$ using a binary \emph{connected score} $C$, defined as: $C(E(x))=1$ if E(x) is $\epsilon$-connected to an instance from $X^{f(E(x))}$, else $C(E(x))=0$.
The value of the parameter $\epsilon$ is of course crucial: it needs to be as small as possible to ensure that the notion is precise enough.

Although assessing this criterion seems complex, it can be noted that its definition resembles the ones used in DBSCAN~\cite{Ester96} clustering algorithm: saying that~$E(x)$ is $\epsilon$-connected to $a_0$ is equivalent to saying~$E(x)$ and $a_0$ both belong to the same DBSCAN cluster with parameters $\epsilon$ and $minPts=2$.

\subsection{Illustrative Examples}

\begin{figure}[t]
\centering     
\includegraphics[width=0.75\columnwidth]
{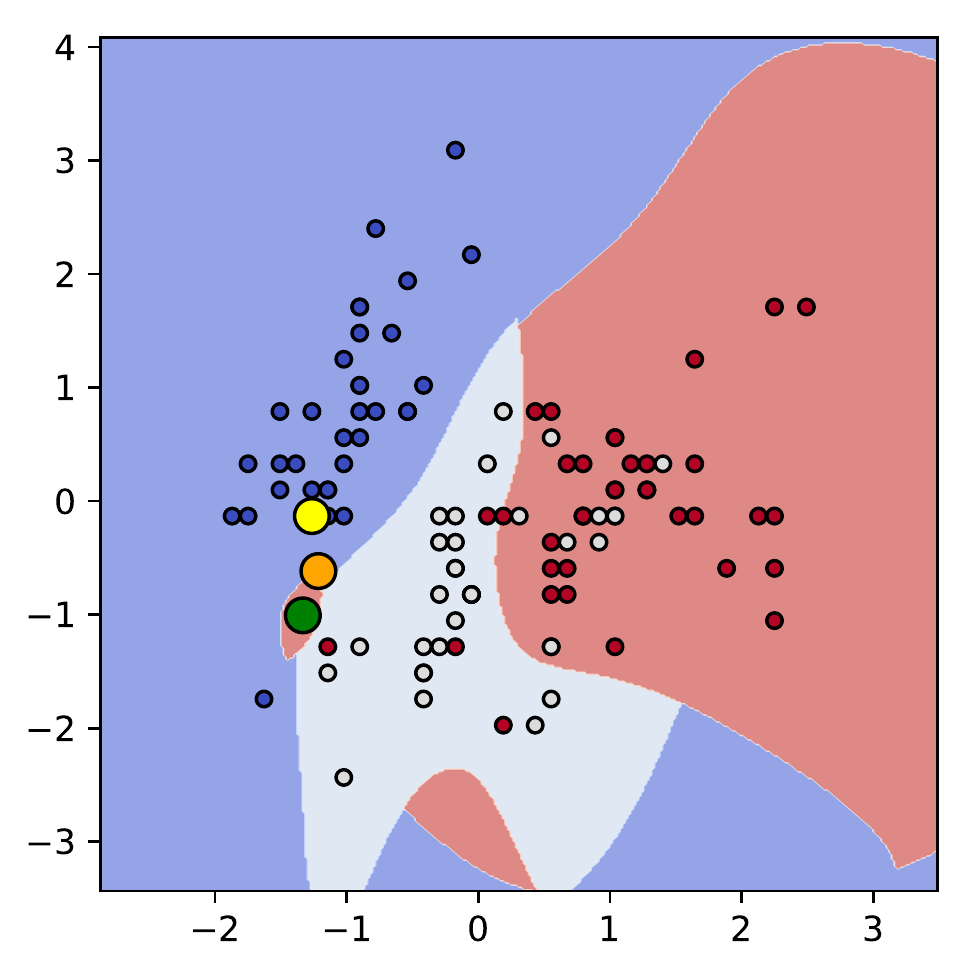}
\caption{Illustration of the intuition behind the Connectedness criterion for two counterfactual explanations generated using HCLS (green) and GS (orange), for a random instance of the iris dataset (yellow).}
\label{fig:illustration-connectedness}
\end{figure}

The same context as in Section~\ref{sec:proximity} is considered: a classifier and two post-hoc counterfactual approaches are used on an instance of a 2D version of the iris dataset. This time, a SVM classifier with a regularization hyperparameter deliberately chosen with a high value ($c=50$) is used (accuracy on test data is 0.73). As illustrated in Figure~\ref{fig:illustration-connectedness}, this strong imposed regularization forces the classifier to create classification regions that do not contain any training instance (two small red regions). In this context, when used for an instance $x$ located nearby (yellow), both HCLS and and GS return counterfactual explanations $E(x)$ that are located in this region. Therefore, they cannot be $\epsilon$-connected to an instance from $X^{f(E(x))}$ and thus both lead to $C(E(x))=0$. Calculating this score for all the instances of this trivial problem shows that $17.8\%$ of the counterfactual examples generated with $E(x)$ are not connected to ground-truth data. This shows that post-hoc approaches are vulnerable to the risk of generating non-connected counterfactual explanations.

\section{Stability}
\label{sec:stability}

\begin{figure}[t]
\centering     
\includegraphics[width=0.95\columnwidth]
{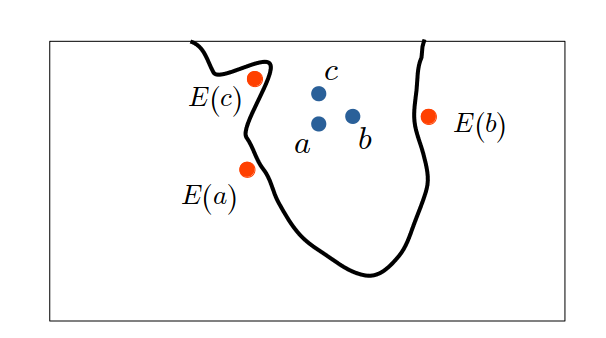}
\caption{Illustration of the intuition behind the Stability notion for three instances $a$, $b$ and $c$.}
\label{fig:illustration-stability}
\end{figure}

Another criterion, mentioned in Section~\ref{sec:background} and sometimes studied for explanations is their stability, sometimes also called robustness~\cite{Alvarez2018}. The idea behind stability is that to be correctly usable, an explanation should be coherent locally, i.e. that its neighbors should have similar explanations. The metrics proposed by~\citet{Alvarez2018} to measure stability is:
$$
\Bar{L}_X(x) =  \underset{x_j \in X \cap \mathcal{B}(x, \epsilon) }{\text{argmax }} \frac{||E(x) - E(x_j)||_2}{||x - x_j||_2}
$$
with $\mathcal{B}(x, \epsilon)$ the hyperball of center $x$ and radius $\epsilon$. While the issue of stability has been observed in some contexts (e.g.~\citet{Smilkov2017} try to smooth gradient-based explanations of deep neural networks by averaging them locally), it still needs refining. This complex notion does not seem indeed to be sufficiently defined, as it is not clear when an observed variation in explanations is the consequence of a lack of robustness of the explainer or of normal variation in the data and in the decision boundary. We illustrate these issues in Figure~\ref{fig:illustration-stability} in a schematic 2-dimensional illustration. Instances $a$, $b$ and $c$ are close to each other but have very different explanations, which may be an issue since it can lead to a lack of trust in the explainer if he/she does not understand these differences. $E(a)$ and $E(b)$ are thus different seemingly because they are close to different classification borders. Another example is instances $a$ and $c$, which have different explanations because of a seemingly lack of robustness of the the classifier, not the explainer. Therefore, a high stability value does not appear to be a consequence of the lack of robustness of the explainer system, but rather a mere information of a variation in the local decision boundary of~$f$.

On the contrary, an exlainer system giving the same explanation, no matter how wrong it is, for every instance of $X$ would achieve perfect stability with respect to the metric~$\Bar{L}_X$. Stability is thus linked to the notion of locality of explanations, studied in~\citet{Laugel2018}. This leads to the question of what stability would be expected of counterfactual explanations, which are by design as local as possible.
Therefore, while the notion of stability is intuitively important to engender trust to the user, it still lacks a proper definition.

\section{Conclusion}
\label{sec:discussion}

In this work we propose the proximity, connectedness and stability as desirable properties to characterize a counterfactual explanation. While other criteria can be proposed to assess correctly the quality of a counterfactual explanation (e.g. d(x, E(x)) is by definition the main quality metric for counterfactuals), they embody reasonable properties a user is probably looking for when using an explainer system. Furthermore, the other goal of this paper is to highlight that these criteria cannot be properly taken into account in a post-hoc context, leading to potential interpretability issues overall.
While this is still on-going work and the proposed metrics need more refining, qualitative results show that they seem to correctly capture the proposed notions.

A natural follow-up question is how to generate counterfactual explanations that satisfy these criteria in the post-hoc context. While no answer can be given with the current form of notions and further research is necessary, it seems that in the mean time, using the training instances, although not always possible in the post-hoc context, remains necessary.

\section*{Aknowledgements}
This work has been done as part of the Joint Research Initiative (JRI) project ”Interpretability for human-friendly machine learning models” funded by the AXA Research Fund.

\bibliography{biblio-counterfactuals.bib}

\begin{thebibliography}{20}
\providecommand{\natexlab}[1]{#1}
\providecommand{\url}[1]{\texttt{#1}}
\expandafter\ifx\csname urlstyle\endcsname\relax
  \providecommand{\doi}[1]{doi: #1}\else
  \providecommand{\doi}{doi: \begingroup \urlstyle{rm}\Url}\fi

\bibitem[Alvarez~Melis \& Jaakkola(2018)Alvarez~Melis and
  Jaakkola]{Alvarez2018}
Alvarez~Melis, D. and Jaakkola, T.
\newblock Towards robust interpretability with self-explaining neural networks.
\newblock \emph{Advances in Neural Information Processing Systems 31}, pp.\
  7786--7795, 2018.

\bibitem[Baehrens et~al.(2010)Baehrens, Schroeter, Harmeling, Hansen, and
  Muller]{Baehrens2010}
Baehrens, D., Schroeter, T., Harmeling, S., Hansen, K., and Muller, K.-R.
\newblock {How to Explain Individual Classification Decisions Motoaki
  Kawanabe}.
\newblock \emph{Journal of Machine Learning Research}, 11:\penalty0 1803--1831,
  2010.

\bibitem[Bottou et~al.(2013)Bottou, Peters, Qui\~{n}onero Candela, Charles,
  Chickering, Portugaly, Ray, Simard, and Snelson]{Bottou2013}
Bottou, L., Peters, J., Qui\~{n}onero Candela, J., Charles, D.~X., Chickering,
  D.~M., Portugaly, E., Ray, D., Simard, P., and Snelson, E.
\newblock Counterfactual reasoning and learning systems: The example of
  computational advertising.
\newblock \emph{Journal of Machine Learning Research}, 14:\penalty0 3207--3260,
  2013.

\bibitem[Breunig et~al.(2000)Breunig, Kriegel, Ng, and Sander]{Breunig2000}
Breunig, M.~M., Kriegel, H.-P., Ng, R.~T., and Sander, J.
\newblock Lof: Identifying density-based local outliers.
\newblock In \emph{Proceedings of the 2000 ACM SIGMOD International Conference
  on Management of Data}, SIGMOD '00, pp.\  93--104, 2000.

\bibitem[Doshi-Velez et~al.(2018)Doshi-Velez, Kortz, Budish, Bavitz, Gershman,
  O\'Brien, Shieber, Waldo, Weinberger, and Wood]{Doshi-Velez2018}
Doshi-Velez, F., Kortz, M., Budish, R., Bavitz, C., Gershman, S.~J., O\'Brien,
  D., Shieber, S., Waldo, J., Weinberger, D., and Wood, A.
\newblock Accountability of {AI} under the law: The role of explanation.
\newblock In \emph{Privacy Law Scholars Conference}, 2018.

\bibitem[Ester et~al.(1996)Ester, Kriegel, Sander, and Xu]{Ester96}
Ester, M., Kriegel, H.-P., Sander, J., and Xu, X.
\newblock A density-based algorithm for discovering clusters in large spatial
  databases with noise.
\newblock In \emph{Proc. of the 2nd Int. Conf. on Knowledge Discovery and Data
  Mining (KDD'96)}, pp.\  226--231, 1996.

\bibitem[Fawzi et~al.(2018)Fawzi, Moosavi-Dezfooli, Frossard, and
  Soatto]{Fawzi2018}
Fawzi, A., Moosavi-Dezfooli, S.-M., Frossard, P., and Soatto, S.
\newblock Empirical study of the topology and geometry of deep networks.
\newblock In \emph{The IEEE Conference on Computer Vision and Pattern
  Recognition (CVPR)}, June 2018.

\bibitem[Guidotti et~al.(2018{\natexlab{a}})Guidotti, Monreale, Ruggieri,
  Pedreschi, Turini, and Giannotti]{Guidotti2018lore}
Guidotti, R., Monreale, A., Ruggieri, S., Pedreschi, D., Turini, F., and
  Giannotti, F.
\newblock Local rule-based explanations of black box decision systems.
\newblock \emph{arXiv preprint 1805.10820}, 2018{\natexlab{a}}.

\bibitem[Guidotti et~al.(2018{\natexlab{b}})Guidotti, Monreale, Ruggieri,
  Turini, Giannotti, and Pedreschi]{Guidotti2018survey}
Guidotti, R., Monreale, A., Ruggieri, S., Turini, F., Giannotti, F., and
  Pedreschi, D.
\newblock A survey of methods for explaining black box models.
\newblock \emph{ACM Computing Surveys (CSUR)}, 51\penalty0 (5):\penalty0 93,
  2018{\natexlab{b}}.

\bibitem[Lash et~al.(2017)Lash, Lin, Street, Robinson, and Ohlmann]{Lash2017}
Lash, M., Lin, Q., Street, N., Robinson, J., and Ohlmann, J.
\newblock Generalized inverse classification.
\newblock In \emph{Proc. of the 2017 {SIAM} Int. Conf. on Data Mining}, pp.\
  162--170, 2017.

\bibitem[Laugel et~al.(2018{\natexlab{a}})Laugel, Lesot, Marsala, Renard, and
  Detyniecki]{Laugel2017inverse}
Laugel, T., Lesot, M.-J., Marsala, C., Renard, X., and Detyniecki, M.
\newblock Comparison-based inverse classification for interpretability in
  machine learning.
\newblock In \emph{Information Processing and Management of Uncertainty in
  Knowledge-Based Systems.}, pp.\  100--111, 2018{\natexlab{a}}.

\bibitem[Laugel et~al.(2018{\natexlab{b}})Laugel, Renard, Lesot, Marsala, and
  Detyniecki]{Laugel2018}
Laugel, T., Renard, X., Lesot, M.-J., Marsala, C., and Detyniecki, M.
\newblock Defining locality for surrogates in post-hoc interpretablity.
\newblock \emph{ICML Workshop on Human Interpretability in Machine Learning
  (WHI 2018)}, 2018{\natexlab{b}}.

\bibitem[Laugel et~al.(2019)Laugel, Lesot, Marsala, Renard, and
  Detyniecki]{Laugel2019}
Laugel, T., Lesot, M.-J., Marsala, C., Renard, X., and Detyniecki, M.
\newblock The dangers of post-hoc interpretability: Unjustified counterfactual
  explanations.
\newblock \emph{to appear in: {IJCAI-19}}, 2019.

\bibitem[Lundberg \& Lee(2017)Lundberg and Lee]{Ljundberg2017}
Lundberg, S.~M. and Lee, S.-I.
\newblock A unified approach to interpreting model predictions.
\newblock In \emph{Advances in Neural Information Processing Systems 30}, pp.\
  4765--4774, 2017.

\bibitem[Martens \& Provost(2014)Martens and Provost]{Martens2014}
Martens, D. and Provost, F.
\newblock Explaining data-driven document classifications.
\newblock \emph{MIS Q.}, 38\penalty0 (1):\penalty0 73--100, 2014.

\bibitem[MC~Grath et~al.(2018)MC~Grath, Costabello, Le~Van, Sweeney, Kamiab,
  Shen, and L\'ecu\'e]{McGrath2018}
MC~Grath, R., Costabello, L., Le~Van, C., Sweeney, P., Kamiab, F., Shen, Z.,
  and L\'ecu\'e, F.
\newblock Interpretable credit application predictions with counterfactual
  explanations.
\newblock \emph{NeurIPS 2018 Workshop on Challenges and Opportunities for {AI}
  in Financial Services: the Impact of Fairness, Explainability, Accuracy and
  Privacy}, 2018.

\bibitem[Rudin(2018)]{Rudin2018}
Rudin, C.
\newblock Please stop explaining black box models for high stakes decisions.
\newblock \emph{NeurIPS Workshop on Critiquing and Correcting Trends in Machine
  Learning}, 2018.

\bibitem[Russell(2019)]{Russell2019}
Russell, C.
\newblock Efficient search for diverse coherent explanations.
\newblock In \emph{Proc. of the Conf. on Fairness, Accountability, and
  Transparency (FAT* '19}, pp.\  20--28, 2019.

\bibitem[Smilkov et~al.(2017)Smilkov, Thorat, Kim, Vi\'egas, and
  Wattenberg]{Smilkov2017}
Smilkov, D., Thorat, N., Kim, B., Vi\'egas, F., and Wattenberg, M.
\newblock Smoothgrad: removing noise by adding noise.
\newblock \emph{arXiv preprint 1706.03825}, 2017.

\bibitem[Wachter et~al.(2018)Wachter, Mittelstadt, and Russell]{Wachter2018}
Wachter, S., Mittelstadt, B., and Russell, C.
\newblock Counterfactual explanations without opening the black box; automated
  decisions and the {GDPR}.
\newblock \emph{Harvard Journal of Law \& Technology}, 31(2):\penalty0
  841--887, 2018.

\end{thebibliography}
\bibliographystyle{icml2019}

\end{document}